\pdfoutput=1

\documentclass[11pt]{article}

\usepackage[table,xcdraw,dvipsnames]{xcolor}

\usepackage{EMNLP2022}

\usepackage{times}
\usepackage{latexsym}

\usepackage[T1]{fontenc}

\usepackage[utf8]{inputenc}

\usepackage{microtype}

\usepackage{inconsolata}

\usepackage{makecell}
\usepackage{graphicx}
\usepackage{bm}
\usepackage{amssymb}
\usepackage[tbtags]{amsmath}
\usepackage{amsfonts}
\usepackage{bbm}

\DeclareMathOperator*{\argmax}{argmax} 

\usepackage{cleveref}
\crefformat{section}{\S#2#1#3}
\Crefname{figure}{Figure}{}
\Crefname{table}{Table}{}
\Crefname{algorithm}{Algorithm}{}
\Crefname{algocf}{Algorithm}{}
\Crefname{equation}{Equation}{}
\crefname{appendix}{Appendix}{}

\usepackage{makecell}
\usepackage{booktabs}
\usepackage{multirow}
\usepackage{tabulary}
\newcolumntype{L}[1]{>{\raggedright\let\newline\\\arraybackslash\hspace{0pt}}m{#1}}
\newcolumntype{C}[1]{>{\centering\let\newline\\\arraybackslash\hspace{0pt}}m{#1}}
\newcolumntype{R}[1]{>{\raggedleft\let\newline\\\arraybackslash\hspace{0pt}}m{#1}}
\usepackage[ruled]{algorithm2e}

\newcommand{\eat}[1]{\ignorespaces}

\usepackage{todonotes}
\newcounter{hoifung}

\newcounter{shikhar}

\newcommand{\kriss}{$\tt KRISS$}
\newcommand{\krissbert}{$\tt KRISSBERT$}
\title{Knowledge-Rich Self-Supervision for Biomedical Entity Linking}

\author{Sheng Zhang\thanks{~~These authors contributed equally to this research.} \quad Hao Cheng$^*$ \quad Shikhar Vashishth$^*$ \quad Cliff Wong \quad Jinfeng Xiao\thanks{~~Work done as an intern at Microsoft Research.} \\
    {\bf Xiaodong Liu} \quad {\bf Tristan Naumann} \quad {\bf Jianfeng Gao} \quad {\bf Hoifung Poon}\\
    Microsoft Research\\
    $^\dagger$University of Illinois at Urbana-Champaign \\}

\begin{document}
\maketitle

\begin{abstract}

Entity linking
faces significant challenges such as prolific variations and prevalent ambiguities, especially in high-value domains with myriad entities. 
Standard classification approaches suffer from the annotation bottleneck and cannot effectively handle unseen entities. 
Zero-shot entity linking has emerged as a promising direction for generalizing to new entities, but it still requires example gold entity mentions during training and canonical descriptions for all entities, both of which are rarely available outside of Wikipedia. 
In this paper, we explore Knowledge-RIch Self-Supervision (\kriss) for biomedical entity linking, by leveraging readily available domain knowledge. 
In training, it generates self-supervised mention examples on unlabeled text using a domain ontology and trains a contextual encoder using contrastive learning. 
For inference, it samples self-supervised mentions as prototypes for each entity and conducts linking by mapping the test mention to the most similar prototype. 
Our approach
can easily incorporate entity descriptions and gold mention labels if available. 
We conducted extensive experiments on seven standard datasets spanning  biomedical literature and clinical notes. Without using any labeled information, our method produces {\krissbert}, a universal entity linker for four million UMLS entities that attains new state of the art, outperforming prior self-supervised methods by as much as 20 absolute points in accuracy. 

\end{abstract}

\section{Introduction}
\vspace{-2mm}

\begin{figure*}[!t]
    \centering
    \includegraphics[width=0.9\textwidth]{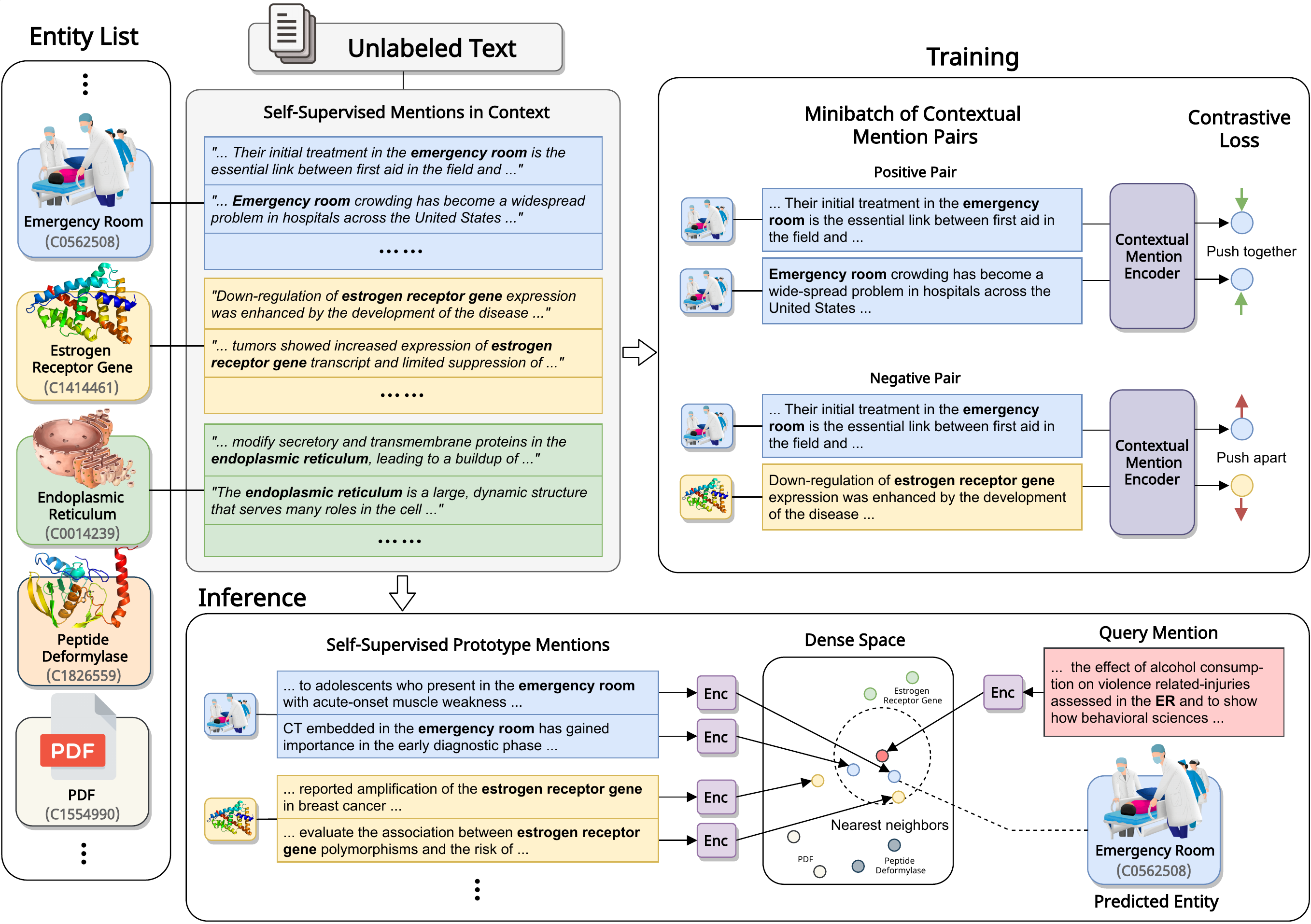}
    \caption{Illustration of knowledge-rich self-supervised entity linking.}
    \vspace{-2mm}
    \label{fig:kriss}
\end{figure*}

Entity linking maps mentions to unique entities in a target knowledge base~\citep{roth-etal-2014-wikification}. It can be viewed as the extreme case of named entity recognition and entity typing, where the category number swells to tens of thousands or even millions. Entity linking is particularly challenging in high-value domains such as biomedicine, where variations and ambiguities abound. 
For instance, depending on the context, {\em ``PDF''} may refer to a gene (Peptide Deformylase, Mitochondrial), 
or file type (Portable Document Format). 
Similarly, {\em ``ER''} could refer to emergency room, the organelle endoplasmicreticulum, or the estrogen receptor gene. 
Moreover, the number of entities in domains such as biomedicine can be very large. The Unified Medical Language System (UMLS), 
a representative ontology for biomedicine, contains over four million entities~\citep{bodenreider2004unified}. 

Standard classification approaches such as MedLinker \cite{medlinker} require example gold mentions for each entity and cannot effectively handle new entities for which there are no labeled examples in training. 
Recently, zero-shot entity linking has emerged as a promising direction for generalizing to unseen entities \cite{logeswaran-etal-2019-zero,wu-etal-2020-scalable},
by learning to encode contextual mentions for similarity comparison against reference entity descriptions.
Existing methods, however, require example gold entity mentions during training, as well as canonical descriptions for all entities. 
While applicable to Wikipedia entities, these methods are hard to generalize to other domains, where such labeled information is rarely available at scale. 

In this paper, we explore \textbf{K}nowledge-\textbf{RI}ch \textbf{S}elf-\textbf{S}upervision  (\kriss) for entity linking by leveraging readily available domain knowledge to compensate for the lack of labeled information (\autoref{fig:kriss}). 
For entity linking, the most relevant knowledge source is the domain ontology. The core of an ontology is the entity list, which specifies the unique identifier and canonical name for each entity and is the prerequisite for entity linking. 
Our method only requires the entity list and unlabeled text, which are readily available in any domain. 

In training, {\kriss} uses the entity list to generate self-supervised mention examples from unlabeled text,
and trains a contextual mention encoder using contrastive learning \cite{gao2014modeling,wu-etal-2020-scalable}, by mapping mentions of the same entity closer. For inference, {\kriss} samples prototypes for each entity from the self-supervised mentions. Given a test mention, {\kriss} finds the most similar prototype and returns the entity it represents. 

Prior methods that leverage domain ontology for entity linking often resort to string matching (against entity names and aliases), making them vulnerable to both variations and ambiguities. 
Recently, a flurry of methods have been proposed to conduct biomedical entity representation learning from synonyms in the ontology, such as BIOSYN~\cite{sung-etal-2020-biomedical}, SapBERT~\cite{liu-etal-2021-self}, and others~\cite{lai-etal-2021-bert-might}. These methods can resolve variations to some extent, but they completely ignore mention contexts and cannot resolve ambiguities. Given an entity mention, they only predict a surface form, rather than a unique entity as required by entity linking (e.g., see footnote 2 in SapBERT~\cite{liu-etal-2021-self}). As we will show in \cref{ssec:sapbert-wrong}, their predicted surface forms are often ambiguous and can't be mapped to a unique entity. Unfortunately, starting from BIOSYN, these papers all adopt an incorrect evaluation method that simply ignores the ambiguity and declares the predicted surface form as correct. Consequently, their reported ``entity linking'' scores are often highly inflated and do not represent true linking performance. In \cref{ssec:sapbert-wrong}, we provide a detailed analysis to illustrate this problem, which we hope would contribute to rectifying this significant evaluation error in future entity linking work.

\eat{
In practice, an ontology may contain additional domain knowledge such as aliases and semantic relations among the entities (e.g., $\tt ISA$ hierarchy). 
Gold mention examples and canonical descriptions may also be available for some entities. 
{\kriss} provides a general framework for incorporating such information.
Aliases can be used to generate additional self-supervised mentions. The $\tt ISA$ hierarchy can be leveraged to aid representation learning for rare entities. 
Gold mentions and entity descriptions can be used as positive examples in contrastive learning, as well as prototypes at inference. 
}

We conduct our study on biomedicine, which serves as a representative high-value domain where prior methods are hard to apply. Among the four million biomedical entities in UMLS, less than 6\% have any description available. Gold mention labels are available for only a tiny fraction of entities. E.g., MedMentions~\cite{mohan2019medmentions}, the largest biomedical entity linking dataset, only covers 35 thousand entities. 

We applied our method to train {\krissbert}, a universal entity linker for all four million biomedical entities in UMLS, using only the entity list in UMLS and unlabeled text in PubMed\footnote{\url{https://pubmed.ncbi.nlm.nih.gov/}}. 
{\krissbert} can also incorporate additional domain knowledge in UMLS such as entity aliases and $\tt ISA$ hierarchy. 
We conducted extensive evaluation on seven standard biomedical entity linking datasets spanning biomedical literature and clinical notes. {\krissbert} demonstrated clear superiority,
outperforming prior state of the art by 10 points in average accuracy and by over 20 points in MedMentions. 

{\krissbert} can be directly applied to lazy learning (\cref{ssec:lazy-learning}) with no additional training, by simply using gold mention examples as prototypes during inference. This universal model already attains comparable results as dataset-specific state-of-the-art supervised methods, each tailored to an individual dataset by limiting entity candidates and using additional supervision sources and more complex methods (e.g., coreference rules and joint inference). 
We released {\krissbert} to facilitate research and applications in biomedical entity linking.\footnote{\url{http://aka.ms/KRISSBERT}}
\section{Related Work}
\vspace{-1mm}

\paragraph{Entity linking} Many applications require mapping mentions to unique entities. E.g., knowing that {\em some drug} can treat {\em some disease} is not very useful, unless we know the specific drug and disease. Entity linking is inherently challenging given the large number of unique entities.
Prior work often adopts a pipeline approach that first narrows entity candidates to a small set (candidate generation) and then learns to classify contexts of the mention and a candidate entity (candidate ranking)~\cite{bunescu-pasca-2006-using,cucerzan-2007-large,ratinov-etal-2011-local}. 
Candidate generation often resorts to string matching or TF-IDF variants (e.g., BM25), which are vulnerable to variations. 
Ranking features are manually engineered or learned via various neural architectures~\cite{he-etal-2013-learning,ganea-hofmann-2017-deep,kolitsas-etal-2018-end}. 
Additionally, entity relations (e.g., concept hierarchy) and joint inference have been explored for improving accuracy~\cite{gupta-etal-2017-entity,murty-etal-2018-hierarchical,cheng-roth-2013-relational,le-titov-2018-improving}. 
These methods are predominantly supervised, and suffer from the scarcity of annotated examples, especially given the large number of entities to cover. 
By contrast, {\krissbert} leverages self-supervision using readily available domain knowledge and unlabeled text, and can effectively resolve variations and ambiguities for millions of entities.

\vspace{-1mm}
\paragraph{Knowledge-rich self supervision} Domain ontology such as UMLS has been applied to self-supervise biomedical named entity recognition \cite{zhang2013unsupervisedBN,almgren-etal-2016-named}. Recently, \citet{sung-etal-2020-biomedical,liu-etal-2021-self} propose SapBERT for mention normalization by conducting contrastive learning over synonyms from UMLS. However, SapBERT completely ignores mention contexts. It can resolve some variations but not ambiguity\footnote{The SapBERT paper states:``{\em In this work, biomedical entity refers to the surface forms of biomedical concepts}''. As aforementioned, many surface forms of biomedical entities are highly ambiguous (e.g., ``{\em PDF}'', ``{\em ER}'').}. 
By contrast, we apply contrastive learning on mention contexts, and leverage unlabeled text to generate self-supervised examples. SapBERT relies on synonyms to learn spelling variations. Our approach can learn with just the canonical name for each entity, as self-supervised mention examples naturally capture contexts where synonymous mentions may appear in. 
\section{Knowledge-Rich Self-Supervision for Entity Linking}

Entity linking grounds textual mentions to unique entities in a given database/dictionary.
Formally, the goal of entity linking is to learn a function ${\tt Link}: (m, T)\rightarrow e$ that maps mention $m$ in the context $T$ to the unique entity $e$.
{\em Self-supervised entity linking} assumes no access to any gold mention examples. 
The knowledge-rich self-supervision setting (\kriss) assumes that only a domain ontology~$\mathcal{O}$ and an unlabeled text corpus~$\mathcal{T}$ are available. 
In particular, we require the availability of an entity list, which specifies for each entity the unique identifier and a canonical name. Entity list is the prerequisite for entity linking, as it provides the targets for linking. Our framework can also incorporate other knowledge in the ontology (\S\ref{ssec:ontology_knowledge}).  

\subsection{Generating Self-Supervision}
\label{ssec:self-supervision}


To generate self-supervised mention examples,
we first compile a list of entity names from preferred terms in UMLS.
We then build a trie from these names (case preserved) to efficiently search them in plain text. When an exact match is found, a fixed-size window around the mention will be returned as context. Some preferred terms are shared by multiple entities. To reduce noise for training and inference, we skip the ambiguous terms. 
We conducted this process on PubMed abstracts and obtained over 1.6 billion mention examples, each of which is uniquely linked to an entity in UMLS. 
The estimated linking accuracy based on random samples is 85\%. 
Note that not all UMLS entities have self-supervised examples, as they have never been mentioned in PubMed. This is not an issue for training as our goal is to learn a general encoder that maps mentions of the same entity closer (\S\ref{ssec:contrasitve_learning}). For inference, the $\tt ISA$ hierarchy in UMLS can be leveraged to compensate for the lack of self-supervised examples (\S\ref{ssec:ontology_knowledge}).

\eat{
First, we collect a list of unambiguous entity names from UMLS. Specifically, for each entity, we attempt to collect names from its preferred terms\footnote{\url{https://www.nlm.nih.gov/research/umls/new_users/online_learning/Meta_004.html}}. However, the preferred terms are not always unambiguous. For example, "\emph{MS}" is used as a preferred term for both \emph{Mass Spectrometry} (C0037813) and \emph{Millisecond} (C0439223).
To reduce ambiguity, we only keep preferred terms that can be mapped to one entity.
Next, we build a trie to search the collected names (with case preserved) in plain text.\footnote{The plain text has not been segmented or tokenized, so the search does not rely on NLP preprocessing tools.} Whenever we find an exact match, we return a fixed-size window around the matched name as context. Since the name can be uniquely mapped to an entity in UMLS, we instantly get a mention example for that entity. 
Finally, we run this automatic process on a large collections of PubMed abstracts, which yields over 1.6 billion mention examples for 534,040 entities.

Our analysis on randomly sampled examples reveals that the estimated accuracy is around 85\%.
Note that we do not generate examples for every entity in UMLS. This is not an issue for training as our goal is to learn an encoder that maps mentions of the same entity closer (\S\ref{ssec:contrasitve_learning}). For inference, we leverage additional knowledge in UMLS to ensure coverage. The details are described in \S\ref{ssec:ontology_knowledge}.
}

\subsection{Contrastive Learning}
\label{ssec:contrasitve_learning}

Given the self-supervised mentions, we train a mention encoder using contrastive learning by mapping mentions of the same entity closer and mentions of different entities farther apart. Specifically,
each mention $m$ is encoded into a contextual vector $\mathbf{c}$ using a Transformer-based encoder~\cite{vaswani2017attention}, with the following input format: 
\begin{center}
\texttt{[CLS]} ctx$_l$ \texttt{[M$_s$]} mention \texttt{[M$_e$]} ctx$_r$ \texttt{[SEP]}
\end{center}
where ctx$_l$ and ctx$_r$ denote the left and right context respectively; \texttt{[M$_s$]} and \texttt{[M$_e$]} are markers indicating the start and end of the mention; \texttt{[CLS]} and \texttt{[SEP]} are special encoding tokens.
The last-layer hidden state of \texttt{[CLS]} is used as the contextual vector $\mathbf{c}$. See \autoref{appendix:encoder} for an illustration.

In a minibatch, we sample $2N$ self-supervised mentions from $N$ entities and encode them into contextual vectors $\{\mathbf{c}_1,...,\mathbf{c}_{2N}\}$, where $\mathbf{c}_{2k-1}$ and $\mathbf{c}_{2k}$ are from the same entity. Given a positive pair $(\mathbf{c}_i,\mathbf{c}_j)$, we treat the other $2(N-1)$ vectors within a minibatch as negative examples, and compute the InfoNCE loss~\cite{oord2018representation} as:
\begin{equation}
\ell_{\mathbf{c}_i,\mathbf{c}_j}=-\log\frac{\exp(\mathbf{c}_i^\top\mathbf{c}_j/\tau)}{\sum_{k=1}^{2N}\mathbbm{1}_{[k\neq i]}\exp(\mathbf{c}_i^\top\mathbf{c}_k/\tau)},
\end{equation}
where $\mathbbm{1}_{[k\neq i]}\in\{0,1\}$ is an indicator function evaluating to 1 iff $k\neq i$ and $\tau$ denotes a temperature parameter.
The final loss is computed across all positive pairs in a minibatch:
\begin{equation}
\label{eq:mm-loss}
\mathcal{L} = \frac{1}{2N}\sum_{k=1}^N[\ell_{\mathbf{c}_{2k-1},\mathbf{c}_{2k}}+\ell_{\mathbf{c}_{2k},\mathbf{c}_{2k-1}}]
\end{equation}

\eat{
A mini-batch comprises $B$ samples from $B/2$ unique entities, with two sampled self-supervised mentions for each: 
$\{m_{b,i}: b\in 1..B/2, i\in\{-1,1\}\}$. 
Each contextual mention is used as the query in turn to compute the InfoNCE loss against others in the batch, with the average loss $\mathcal{L}$ being:
\vspace{-5pt}
\begin{equation*}
\small
\frac{1}{B}\sum_{b,i}-\log\frac{\exp(C(m_{b,i})\cdot C(m_{b,-i})/\tau)}{\sum\limits_{(b',i')\neq (b,i)}\exp(C(m_{b,i})\cdot C(m_{b',i'})/\tau)}
\end{equation*}
where $C(\cdot)$ refers to the contextual mention encoder and $\tau$ is the temperature. We use a transformer-based model with entity markers added around each mention, and return the top \texttt{[CLS]} encoding (Appendix \autoref{fig:encoder}). 
}

\eat{
Specifically, a contextual mention example is given as
\begin{center}
\texttt{[CLS]} ctx$_l$ \texttt{[M$_s$]} mention \texttt{[M$_e$]} ctx$_r$ \texttt{[SEP]},
\end{center}
where ctx$_l$/ctx$_r$ are the left/right contexts, \texttt{[M$_s$]}/\texttt{[M$_e$]} are the start/end mention markers, and \texttt{[CLS]}/\texttt{[SEP]} are special tokens. 
After processing by the encoder, the \texttt{[CLS]} vector in the last transformer layer will be used as the contextual encoding $C(m)$. 
}

\subsection{Mention Masking and Replacement}
\label{ssec:mention}

Skipping ambiguous names improves the quality of mention examples (\S\ref{ssec:self-supervision}), but models trained with such self-supervision tend to over-index on surface matching, limiting generalizability.
To overcome this, we propose two strategies to augment alternative views of the encoder input during training:
 

\noindent\textbf{Mention Masking}
With a probability $p_{\text{mask}}$, we mask the mention using \texttt{[MASK]}, which regularizes the model from lexical memorization and encourages it to leverage cues from surrounding context.

\noindent\textbf{Mention Replacement} With a probability $p_\text{replace}$, the mention is replaced with its synonym in UMLS while the context is kept unchanged. This yields a new mention of the same entity, encouraging the model to generalize across entity variations.

\subsection{Linking with Self-Supervised Prototypes}
\label{ssec:inference}

At test time, for each entity $e$ in the entity list $\mathcal{E}$ compiled in \cref{ssec:self-supervision}, we sample a small set of self-supervised mentions as reference prototypes, denoted as ${\tt Proto}(e)$. 
Given a test/query mention $m_q$, we return the entity $e$ with the most similar prototype $m_p$ based on the self-supervised encoding: 
\vspace{-5pt}
\begin{equation}
\label{eq:link}
{\tt Link}(m_q)=\argmax_{e\in\mathcal{E}}\max_{m_p\in {\tt Proto}(e)}\mathbf{c}_q^\top\mathbf{c}_p 
\end{equation}

\eat{

For inference, we pre-compute vectors for all entity-centric references and update the linking function defined in Equation (\ref{eq:link}) as below:
\begin{equation}
\label{eq:link2}
{\tt Link}(m_t)=\argmax_{e\in\mathcal{E}}\max_{m_i\in {\tt Proto}(e)}(\mathbf{c}^\top_t\mathbf{c}_i + \mathbf{c}_t^\top\mathbf{r}_e)
\end{equation}
}

\eat{
\begin{equation}
\label{eq:link}
{\tt Link}(m_i)=\argmax_{e\in\text{KB}}\max_{m_j\in {\tt Proto}(e)}\mathbf{c}^\top_i\mathbf{c}_j 
\end{equation}
Each prototype is uniquely mapped to an entity. Therefore, our approach directly returns an unambiguous entity ID.
This differs from \citet{sung-etal-2020-biomedical,liu-etal-2021-self}, which return an entity name that can be used by multiple entities and needs further disambiguation.
Finally, since prototypes are encoded independently, we pre-compute their contextual vectors and leverage nearest neighbor search tools such as FAISS \cite{johnson2019billion} for large-scale linking.
}
\vspace{-5pt}
For efficient linking, we pre-compute the contextual vectors of all reference prototypes and leverage fast nearest neighbor search tool 
that can scale to millions of entities~\cite{johnson2019billion}. 

\subsection{Incorporating Additional Knowledge}
\label{ssec:ontology_knowledge}

Our self-supervised entity linking formulation can easily incorporate other knowledge available in an ontology, either by generating additional mention examples from unlabeled text, or by creating special entity-centric examples, which can be used both for learning and inference.
This is especially important for entities without self-supervised mentions from PubMed (\S\ref{ssec:self-supervision}). 

\noindent\textbf{Aliases} Ontology often includes aliases for some entities. The alias lists are generally incomplete and aliases such as acronyms are highly ambiguous. So they can't be used as a definitive source for candidate generation. However, aliases can be used in {\kriss} to generate additional self-supervised mentions from unlabeled text, just like the preferred terms. To reduce noise, we similarly skip ambiguous aliases shared by multiple entities. 

\noindent\textbf{Semantic hierarchy} Ontology often organizes entities in a hierarchy via $\tt ISA$ relationship among entities.
For instance, in UMLS, the ER gene is assigned a {\em Semantic Tree Number} (A1.2.3.5), which specifies the $\tt ISA$ path from root to its entity type ({\em Gene or Genome}). 
For each entity in UMLS, we concatenate its semantic tree number (stn), entity type, as well as aliases to generate an \emph{entity-centric reference} in the following form:
\begin{center}
\texttt{[CLS]} stn \texttt{[SEP]} type \texttt{[SEP]} aliases \texttt{[SEP]}
\end{center}
We introduce a separate encoder to compute the vector representation $\mathbf{r}_e$ from the last-layer hidden state of {\tt[CLS]} for entity $e$. 
For learning, besides the contextual vectors $\{\mathbf{c}_1,...,\mathbf{c}_{2N}\}$ for $N$ entities, a minibatch includes $N$ entity-centric renferences $\{\mathbf{r}_{e_1},...,\mathbf{r}_{e_N}\}$.
Given a positive pair $(\mathbf{c}_i,\mathbf{r}_{e_j})$, we treat the other $N-1$ entity-centric references as negatives and compute the InfoNCE loss:
\begin{equation}
\ell_{\mathbf{c}_i,\mathbf{r}_{e_j}}=-\log\frac{\exp(\mathbf{c}_i^\top\mathbf{r}_{e_j}/\pi)}{\sum_{k=1}^{N}\exp(\mathbf{c}_i^\top\mathbf{r}_{e_k}/\pi)},
\end{equation}
where $\pi$ is a temperature parameter. The final loss between mentions and entity-centric references is computed across all positive pairs in a minibatch:
\begin{equation}
\mathcal{L}' = \frac{1}{2N}\sum_{k=1}^N[\ell_{\mathbf{c}_{2k-1},\mathbf{r}_{e_k}}+\ell_{\mathbf{c}_{2k},\mathbf{r}_{e_k}}]
\end{equation}
We jointly optimize two contrastive losses $\alpha\mathcal{L} +\beta\mathcal{L}'$, with weights $\alpha$ and $\beta$.
\eat{
: 
\vspace{-1mm}
\begin{center}
\texttt{[CLS]} STN \texttt{[SEP]} ET \texttt{[SEP]} aliases \texttt{[SEP]}.
\end{center}
}
For inference, we include entity-centric references in ${\tt Link}(m_q)$ as:
\vspace{-1mm}
\begin{equation}
\label{eq:link2}
{\tt Link}(m_q)=\argmax_{e\in\mathcal{E}}\max_{m_p\in {\tt Proto}(e)}\mathbf{c}^\top_q(\mathbf{c}_p + \mathbf{r}_e)
\end{equation}

\vspace{-3mm}
\paragraph{Entity description} For a small fraction of common entities, manually written descriptions may be available. In UMLS, less than 6\% of entities have description, so they can't be used as the main source for contrastive learning and linking. Still, the information may be useful and can be incorporated in {\kriss} by appending it to the entity-centric reference (separated by \texttt{[SEP]}).

\subsection{Cross-Attention Candidate Re-Ranking}

Inspired by \citet{logeswaran-etal-2019-zero,wu-etal-2020-scalable}, we further improve the linking accuracy by learning to re-rank the top $K$ candidates via a \emph{cross-attention} encoder.
The input concatenates the mention and candidate representations (with the second {\tt [CLS]} removed). 
A linear layer is applied to the top {\tt [CLS]} encoding to compute the re-ranking score. 
The training data is generated by pairing self-supervised mentions 
with top $K$ candidates based on ${\tt Link}(m_t)$.
We learn the encoder using a cross-entropy loss that maximizes the re-ranking score for the correct entity.


\subsection{Lazy Learning}
\label{ssec:lazy-learning}

{\kriss} does not require labeled information in training or inference.
However, if labeled examples are available, {\kriss} can directly use them, with zero additional training, as in \emph{lazy learning}~\cite{Wettschereck2004ARA}. In this case, gold mention examples from target training data are used as mention prototypes for linking, augmenting the self-supervised ones. 
{\kriss} can also use labeled examples to fine-tune the self-supervised model, 
we leave it to future work.

\begin{table}[!t]
\small
\centering
\begin{tabular}{@{}lrrr@{}}
\toprule
 & \textbf{Mentions} & \textbf{Entities} & \textbf{Domain Entities} \\ \midrule
NCBI & 6,892 & 790 & 16,317 \\
BC5CDR-d & 5,818 & 1,076 & 16,317 \\
BC5CDR-c & 4,409 & 1,164 & 233,632 \\
ShARe & 17,809 & 1,866 & 82,763 \\
N2C2 & 13,609 & 3,791 & 423,670 \\
MM (full) & 352,496 & 34,724 & 3,416,210 \\
MM (st21pv) & 203,282 & 25,419 & 2,325,023 \\ \bottomrule
\end{tabular}
\vspace{-3mm}
\caption{Summary of entity linking datasets used in our evaluation. MM refers to MedMentions; st21pv refers to the subset with 21 most common semantic types. Domain entities refer to candidates in the UMLS sub-domains (e.g., disease) considered in the dataset.}
\label{tab:data-stats}
\end{table}

\begin{table*}[!ht]
\small
\centering
\resizebox{\textwidth}{!}{
\begin{tabular}{llllllllc}
\Xhline{1pt}
\\[-8pt]
 & NCBI & BC5CDR-d & BC5CDR-c & ShARe & N2C2 & MM (full) & MM (st21pv) & Mean \\ \Xhline{0.5pt}
\\[-8pt]
QuickUMLS & 39.7 & \quad 47.5 & \quad 34.9 & 42.1 & 29.8 & \quad 12.1 & \quad 20.0 & 32.3 \\
BLINK & 49.0 & \quad 48.7 & \quad 52.0 & 32.8 & 25.1 & \quad 13.9 & \quad 19.4 & 34.4 \\
SapBERT$^\dagger$ & 63.0 & \quad 83.6 & \quad 96.2 & 80.4 & 59.7 & \quad 37.6 & \quad 44.2 & 66.4 \\
\textbf{\krissbert}{\scriptsize~(self-supervised)} & \textbf{83.2}{\tiny $\pm$0.5} & \quad \textbf{85.5}{\tiny $\pm$0.2} & \quad \textbf{96.5}{\tiny $\pm$0.1} & \textbf{84.0}{\tiny $\pm$0.1} & \textbf{67.8}{\tiny $\pm$0.1} & \quad \textbf{61.4}{\tiny $\pm$0.1} & \quad \textbf{63.5}{\tiny $\pm$0.1} & \textbf{77.4} \\  
\Xhline{0.5pt}
MedLinker & \cellcolor[HTML]{DAE8FC}50.5 & \quad \cellcolor[HTML]{DAE8FC}62.0 & \quad \cellcolor[HTML]{DAE8FC}80.5 & \cellcolor[HTML]{DAE8FC}56.8 & \cellcolor[HTML]{DAE8FC}37.6 & \quad \cellcolor[HTML]{DAE8FC}32.9 & \quad \cellcolor[HTML]{DAE8FC}57.6 & \cellcolor[HTML]{DAE8FC}54.0 \\
ScispaCy & \cellcolor[HTML]{DAE8FC}66.8 & \quad \cellcolor[HTML]{DAE8FC}64.0 & \quad \cellcolor[HTML]{DAE8FC}85.3 & \cellcolor[HTML]{DAE8FC}66.6 & \cellcolor[HTML]{DAE8FC}54.6 & \quad \cellcolor[HTML]{DAE8FC}53.1 & \quad \cellcolor[HTML]{DAE8FC}52.9 & \cellcolor[HTML]{DAE8FC}63.3 \\
{\krissbert}{\scriptsize~(supervised only)} & \cellcolor[HTML]{DAE8FC}76.9{\tiny $\pm$0.9} & \quad \cellcolor[HTML]{DAE8FC}85.5{\tiny $\pm$0.7} & \quad \cellcolor[HTML]{DAE8FC}93.8{\tiny $\pm$0.3} & \cellcolor[HTML]{DAE8FC}53.9{\tiny $\pm$0.4} & \cellcolor[HTML]{DAE8FC}29.2{\tiny $\pm$1.2} & \quad \cellcolor[HTML]{DAE8FC}60.7{\tiny $\pm$0.3} & \quad \cellcolor[HTML]{DAE8FC}63.7{\tiny $\pm$0.4} & \cellcolor[HTML]{DAE8FC}66.2 \\
{\textbf{\krissbert}}{\scriptsize~(lazy supervised)} & \cellcolor[HTML]{DAE8FC}\textbf{89.9}{\tiny $\pm$0.1} & \cellcolor[HTML]{DAE8FC}\quad \textbf{90.7}{\tiny $\pm$0.1} & \cellcolor[HTML]{DAE8FC}\quad \textbf{96.9}{\tiny $\pm$0.1} & \cellcolor[HTML]{DAE8FC}\textbf{90.4}{\tiny $\pm$0.1} & \cellcolor[HTML]{DAE8FC}\textbf{80.2}{\tiny $\pm$0.1} & \quad \cellcolor[HTML]{DAE8FC}\textbf{70.7}{\tiny $\pm$0.1} & \quad \cellcolor[HTML]{DAE8FC}\textbf{70.6}{\tiny $\pm$0.1} &  \cellcolor[HTML]{DAE8FC}\textbf{84.2} \\ \Xhline{1pt}
\end{tabular}
}
\vspace{-3mm}
\caption{
Comparison of test accuracy on standard entity linking datasets.
Top four systems only use UMLS and unlabeled text. MedLinker and ScispaCy use MedMentions labeled examples for supervision. {\krissbert}{\small~(self-supervised)} uses self-supervised mentions for learning and linking, whereas {\krissbert}{\small~(supervised only)} uses training-set mentions instead. {\krissbert}{\small~(lazy supervised)} augments {\krissbert}{\small~(self-supervised)} with training-set mentions for linking, as in lazy learning (\cref{ssec:lazy-learning}). $^\dagger${\bf SapBERT results are different from reported in \citet{liu-etal-2021-self}. We explain the difference in \cref{ssec:sapbert-wrong}.}
}
\label{tab:main-results}
\end{table*}

\section{Experiments}


\subsection{Entity Linking Benchmark}

%
We consider seven standard entity linking datasets, spanning biomedical literature and clinical notes. See \autoref{tab:data-stats} for a summary. 
In particular, MedMentions~(MM)~\cite{mohan2019medmentions} is the largest and most comprehensive dataset for biomedical entity linking, covering diverse UMLS entities (including all entity types in other datasets). See \autoref{appendix:dataset} for details. 
Training and development sets are not used in any way during self-supervised learning.
Only test sets are used to evaluate self-supervised entity linking. 
we assume that gold mention boundaries are given and focus on evaluating linking accuracy. Given a test mention, the system needs to return the correct entity unique identifier to be considered as correct, as required by entity linking. 

\subsection{Implementation Details}

For unlabeled text, we use the same corpus as in \citet{pubmedbert}, comprising 14 million PubMed abstracts. 
For domain ontology, we use UMLS 2017AA Active, 
 containing 3.47 million entities. 


We use a self-supervised dev set to choose hyperparameters. 
For self-supervised mentions, a  mention-centered window of 64 tokens is used as context. We sample three mentions per entity for training, and sixteen as prototypes at test time. 
The encoders for mentions and entity-centric references are initialized with PubMedBERT~\cite{pubmedbert}. For learning, we use Adam with batch size 512, learning rate 10$^{-5}$, dropout rate 0.1, and both $p_\text{mask}$ and $p_\text{replace}$ 0.2. For simplicity, we set temperatures $\tau, \pi$ to 1.0, and loss weights $\alpha,\beta$ to 0.5. 
Training takes 3 hours on 4 NVIDIA V100 GPUs.
We update parameters in all layers and denote the end model as \krissbert. 
At test time, 
we use FAISS~\cite{johnson2019billion} with IndexFlatIP to obtain the top 100 prototypes for re-ranking.

\subsection{Baseline Systems}

We conduct head-to-head comparison against five baseline systems, including popular tools and prior state-of-the-art methods: 
QuickUMLS~\cite{soldaini2016quickumls}, BLINK~\cite{wu-etal-2020-scalable}, SapBERT~\cite{liu-etal-2021-self}, MedLinker~\cite{medlinker}, ScispaCy~\cite{neumann-etal-2019-scispacy}. See \autoref{appendix:baseline} for details.

\subsection{Main Results}
\label{ssec:main-results}

\autoref{tab:main-results} shows the main results.
{\krissbert} results are averaged over three runs with different random seeds.
As expected, QuickUMLS provides a reasonable dictionary-based baseline but can't effectively handle variations and ambiguities. 
BLINK attained promising results in the Wikipedia domain, but performed poorly in biomedical entity linking, due to the scarcity of available entity descriptions.
SapBERT performed well on largely unambiguous entity types such as chemicals/drugs but faltered in more challenging datasets such as MedMentions.
By contrast, {\krissbert} performed substantially better across the board, establishing new state of the art in self-supervised biomedical entity linking, outperforming prior best systems by 10 points in average and by over 20 points in MedMentions. 
The SapBERT results are different from \citet{liu-etal-2021-self}; we explain the difference in \Cref{ssec:sapbert-wrong}.

By leveraging knowledge-rich self-supervision, {\krissbert} even substantially outperformed supervised entity linkers such as MedLinker and ScispaCy, which used MedMention training data, gaining over 10-20 absolute points in average.  

Self-supervised {\krissbert} also outperforms {\krissbert}{~(supervised only)}. 
It is particularly remarkable as {\krissbert}{~(self-supervised)} learns a \emph{single, unified} model for over three million UMLS entities, whereas {\krissbert}{~(supervised only)} learns \emph{separate} supervised models that tailor to individual datasets. 
This seemingly counter-intuitive result can be explained by {\em the unreasonable effectiveness of data}~\cite{halevy-et-al-2009-data-effect}. 
Knowledge-rich self-supervision produces a large dataset comprising diverse entity and mention examples. Despite the inherent noise, it confers significant advantage over supervised learning with small training data. This manifests most prominently in small clinical datasets like ShARe and N2C2.

\subsection{Why the Entity Linking Scores Reported in the SapBERT Paper Are Incorrect?}
\label{ssec:sapbert-wrong}

The SapBERT paper~\cite{liu-etal-2021-self} reported substantially higher scores than that in \autoref{tab:main-results}. Unfortunately, this stems from a significant error in their evaluation method, as inherited from BIOSYN~\cite{sung-etal-2020-biomedical} and widely adopted in subsequent work (e.g., \citealp{lai-etal-2021-bert-might}). Here, we conduct a detailed analysis using SapBERT~\cite{liu-etal-2021-self} as the representative example. 

The problem can be immediately discerned by first principle. SapBERT completely ignores the context of an entity mention (e.g., see Footnote 3 and Formal Definition in Section 2 in \citealp{liu-etal-2021-self}). Given an ambiguous mention, there is no way such methods can resolve the ambiguity. Instead, these methods would merely produce a surface form (Footnote 2 in \citealp{liu-etal-2021-self}). If the surface form matches multiple entities in name or alias, these methods can't predict a unique entity as required by entity linking. Unfortunately, such an ambiguous prediction is considered correct by their evaluation, as long as the gold entity is one of the matching entities.\footnote{See the code at \url{https://github.com/cambridgeltl/sapbert/blob/main/evaluation/utils.py\#L42}} 

\begin{table}[!t]
\small
\centering
\begin{tabular}{@{}L{0.49\textwidth}@{}}
\toprule
\textbf{Mention}: ``\emph{... Hence, we aimed to find drug targets using the 2DE / \underline{\textbf{MS}} proteomics study of a dexamethasone ...}'' \\
\textbf{SapBERT prediction}: \textcolor{Maroon}{surface form \textsc{ms}, which is shared by multiple entities, such as Master of Science (C1513009), Mass Spectrometry (C0037813), etc.} \\
\textbf{\krissbert~prediction}: \textcolor{NavyBlue}{Mass Spectrometry (C0037813)} \\
\textbf{\krissbert~predicted prototype}: \textcolor{NavyBlue}{\emph{``... \underline{\textbf{mass spectrometry}} is a widely used technique for enrichment and sequencing of phosphopeptides ...''}}\\ \midrule
\textbf{Example}: ``\emph{... every patient followed up accordingly within ten days of \underline{\textbf{discharge}} ...}'' \\
\textbf{SapBERT prediction}: \textcolor{Maroon}{surface form \textsc{discharge}, which is shared by multiple entities, such as Discharge, Body Substance, Sample (C0600083), Patient Discharge (C0030685), etc.} \\
\textbf{\krissbert~prediction}: \textcolor{NavyBlue}{Patient Discharge (C0030685)} \\
\textbf{\krissbert~predicted prototype}: \textcolor{NavyBlue}{\emph{``Performance of the Hendrich Fall Risk Model II in Patients \underline{\textbf{Discharged}} from Rehabilitation Wards ...''}}\\ \midrule
\textbf{Example}: ``\emph{... 5 days of oral prednisone in treatment of adults with \underline{\textbf{mild}} to moderate asthma exacerbations ...}'' \\
\textbf{SapBERT prediction}: \textcolor{Maroon}{surface form \textsc{mild}, which is shared by multiple entities, such as Mild Severity of Illness Code (C1547225), Mild Adverse Event (C1513302).} \\
\textbf{\krissbert~prediction}: \textcolor{NavyBlue}{Mild asthma (C0581124)} \\
\textbf{\krissbert~predicted prototype}: \textcolor{NavyBlue}{\emph{``\underline{\textbf{Mild asthma}}\,exacerbations in a group of children with cough as a dominant symptom ...''}}\\ \midrule
\textbf{Example}: ``\emph{... in patients with thyroid nodules evaluated as Bethesda Category III ( \underline{\textbf{AUS}} / FLUS) in cytology ...}'' \\
\textbf{SapBERT prediction}: \textcolor{Maroon}{surface form \textsc{aus}, which is used by Australia (C0004340).} \\
\textbf{\krissbert~prediction}: \textcolor{NavyBlue}{Atypical cells of undetermined significance (C0522580)} \\
\textbf{\krissbert~predicted prototype}: \textcolor{NavyBlue}{\emph{``\underline{\textbf{Atypia of undetermined significance}} (AUS) or follicular lesion of undetermined significance (FLUS), as stated by The Bethesda System for Reporting Thyroid Cytopathology ...''}}\\
\bottomrule
\end{tabular}
\vspace{-3mm}
\caption{
Examples of ambiguous mentions: SapBERT struggles whereas {\krissbert} predicts correctly.
}
\label{tab:improvement}
\end{table}

\autoref{tab:improvement} shows examples of such ambiguous cases. E.g., given the mention ``MS'', without the context SapBERT has no way to resolve its ambiguity. Instead, it simply returns a verbatim surface form ``MS'', which can be mapped to many UMLS entities. Following BIOSYN, SapBERT evaluation would simply considers this as correct, as one of the matching entities is the gold entity Mass Spectrometry (C0037813). However, this obviously does not reflect the true linking performance for SapBERT, as it can't distinguish it from other equally matching entities such as such as Master of Science (C1513009) and Montserrat Island (C0026514).

\begin{table}[!t]
\small
\centering
\begin{tabular}{@{}lccc@{}}
\toprule
& Mention As-is & SapBERT & \krissbert \\ \midrule
NCBI & 76.9 & 92.0 & 91.3 \\
BC5CDR-d & 83.4 & 93.8 & 92.8 \\
BC5CDR-c & 92.3 & 96.5 & 97.2 \\
ShARe & 74.5 & 85.6 & 87.3 \\
N2C2 & 61.2 & 67.9 & 76.1 \\
MM (full) & 47.1 & 52.2 & 71.3 \\
MM (st21pv) & 48.3 & 53.8 & 72.2 \\ 
\midrule
Mean & 69.1 & 77.4 & 84.0 \\
\bottomrule
\end{tabular}
\caption{Accuracy comparison based on the evaluation metric used by \citet{liu-etal-2021-self}.}
\label{tab:use-sapbert-eval}
\end{table}

\begin{table}[!t]
\small
\centering
\begin{tabular}{@{}lccc@{}}
\toprule
& Ambiguous(\%) & SapBERT & {\krissbert} \\ \midrule
NCBI & 43.2 & 57.1 & 64.5 \\
BC5CDR-d & 30.7 & 63.9  & 64.5 \\
BC5CDR-c & 11.5 & 76.4 & 76.5 \\
ShARe & 48.5 & 67.5 & 72.4 \\
N2C2 & 67.5 & 50.7 & 58.2 \\
MM (full) & 67.8 & 24.8 & 48.9 \\
MM (st21pv) & 69.4 & 29.6 & 52.5 \\ \bottomrule
\end{tabular}
\vspace{-3mm}
\caption{Accuracy comparison on ambiguous cases.}
\label{tab:ambiguous}
\end{table}

\begin{table}[!t]
\small
\centering
\begin{tabular}{@{}lcl@{}}
\toprule
 & \begin{tabular}[c]{@{}c@{}}{\krissbert}\\ (lazy supervised)\end{tabular} & \begin{tabular}[c]{@{}c@{}} Supervised \\ State of the Art\end{tabular} \\ \midrule
NCBI & 89.9 & 89.1~{\scriptsize\cite{ji2020bert}} \\
BC5CDR & 93.7 & 91.3~{\scriptsize\cite{angell-etal-2021-clustering}} \\
ShARe & 90.4 & 91.1~{\scriptsize\cite{ji2020bert}} \\
N2C2 & 80.2 & 81.6~{\scriptsize\cite{xu2020unified}} \\
MM (full) & 70.7 & 45.3$^\dagger${\scriptsize\cite{mohan2019medmentions}}\\ 
MM (st21pv) & 70.6 & 74.1~{\scriptsize\cite{angell-etal-2021-clustering}} \\ \bottomrule 
\end{tabular}
\vspace{-3mm}
\caption{
Comparison of test accuracy of {\krissbert} with lazy learning (\cref{ssec:lazy-learning}) and supervised state of the art.
$^\dagger$Prior work generally avoids evaluating on the full MM dataset; we can only find one published result which combines boundary detection and linking.
}
\label{tab:supervised}
\end{table}

\begin{table*}[!t]

\resizebox{\textwidth}{!}{
\small
\centering
\begin{tabular}{lcccccccc}
\Xhline{1pt}
\\[-8pt]
 & \multicolumn{1}{l}{NCBI} & BC5CDR-d & BC5CDR-c & ShARe & N2C2 & MM (full) & MM (st21pv) & \multicolumn{1}{c}{{\bf Mean}} \\ \Xhline{0.5pt}
\\[-8pt]
\textbf{\krissbert} & \textbf{83.2} & \textbf{85.5} & \textbf{96.5} & \textbf{84.0} & \textbf{67.8} & \textbf{61.4} & \textbf{63.5} & \textbf{77.4} \\
\quad $-$ cross-attention re-ranking & 82.8 & 85.0 & 95.1 & 83.4 & 65.0 & 59.4 & 61.3 & 76.0 \\
\quad $-$ mention pair contrast & 77.9 & 82.2 & 93.3 & 75.0 & 56.3 & 47.8 & 49.9 & 68.9 \\
\quad $-$ aliases & 83.2 & 85.2 & 96.4 & 84.0 & 67.7 & 61.0 & 63.2 & 77.2 \\
\quad $-$ semantic hierarchy & 82.7 & 85.1 & 96.4 & 83.0 & 65.7 & 59.0 & 61.5 & 76.3 \\
\quad $-$ entity description & 83.1 & 85.4 & 96.3 & 84.0 & 67.8 & 61.2 & 63.4 & 77.3 \\
\quad
Initialize w. BERT & 79.3 & 80.6 & 94.4 & 74.5 & 58.4 & 53.9 & 55.3 & 70.9 \\ 
\Xhline{1pt}
\end{tabular}
}
\vspace{-3mm}
\caption{
Ablation study of {\krissbert} on the impact of knowledge components and domain-specific pretraining.
}
\label{tab:ablation}
\end{table*}

\begin{figure*}[!t]
    \centering
    \includegraphics[width=0.95\textwidth]{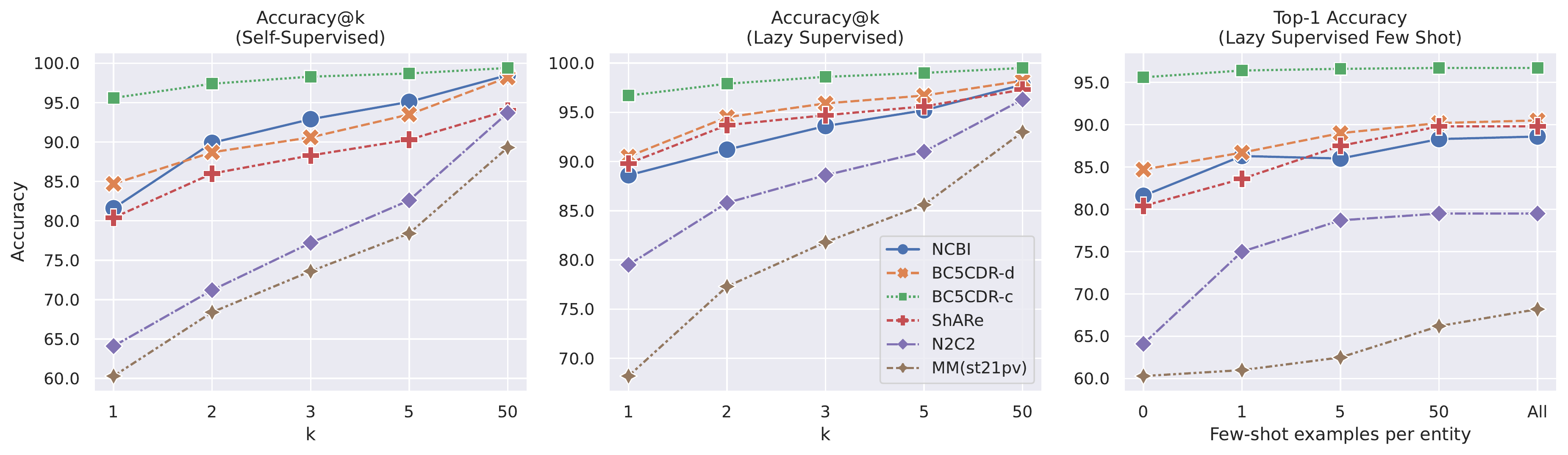}
    \vspace{-3mm}
    \caption{
    Test accuracy (oracle) with top $K$ predictions shows that improving ranking has the potential to yield large gains. Few-shot learning results are averaged over three runs.
    }
    \label{fig:improve}
\end{figure*}

Even if we adopt this incorrect evaluation method, {\krissbert} still substantially outperforms SapBERT, especially on the largest and most challenging MedMention dataset (see \autoref{tab:use-sapbert-eval}). The gain stems from cases when the gold entities have no official aliases matching the surface form predicted by SapBERT, whereas {\krissbert} can still match the gold entity based on context (e.g., see the last two examples in \autoref{tab:improvement}). 
We also evaluated the trivial baseline that returned the mention as is and found that SapBERT often does not outperform it by much, especially on the most representative MedMention dataset. 
Interestingly, under this inflated evaluation, {\krissbert} appears to slightly underperform SapBERT in the relatively easy datasets NCBI and BC5CDR-d (both about diseases). We found that, in rare occasions, the context may lead {\krissbert} to predict a more fine-grained concept (see \autoref{appendix:errors}).



As shown in \autoref{tab:ambiguous}, ambiguous mentions\footnote{We consider a mention as ambiguous if it can't be matched to a unique entity as is.} abound, especially in more diverse and realistic datasets such as N2C2 and MedMentions. The SapBERT paper's evaluation thus reflects the oracle score (assuming that the right entity is always chosen out of multiple candidates), rather than true linking performance. For more realistic assessment, if SapBERT returns multiple entities, a random one would be chosen for evaluation, as in \cref{ssec:main-results}. 
Not surprisingly, {\krissbert} substantially outperforms SapBERT in the ambiguous cases, but still has much room for growth.

\subsection{Lazy Supervised Entity Linking}

{\krissbert} can make good use of labeled data when available. Even lazy learning (\S\ref{ssec:lazy-learning}) yields results comparable to supervised state of the art, as shown in \Cref{tab:supervised}. 
Note that \krissbert{~(lazy supervised)} is based on a single task-agnostic model (\krissbert{~(self-supervised)}), and simply uses corresponding training set examples as prototypes for linking in a zero-shot fashion. By contrast, prior supervised state-of-the-art results were attained using separate models that tailored to individual datasets. They may use additional supervision such as coreference and joint inference~\cite{angell-etal-2021-clustering}, which can be incorporated into \krissbert.

\subsection{Ablation Studies}

In \autoref{tab:ablation}, we conduct a series of ablation studies to understand the impact of domain knowledge and model choices.
Deep cross attention between query mentions and candidates produces consistent gains.
The mention pair contrastive loss $\mathcal{L}$ (\Cref{ssec:contrasitve_learning}) is fundamental for self-supervised learning, whereas additional domain knowledge such as entity descriptions and semantic hierarchy offer incremental gains. 
Domain-specific pretraining \citep[PubMedBERT;][]{pubmedbert} offers a substantial advantage for biomedical entity linking, gaining 6.5 points on average over BERT initialization.

\subsection{Discussion}

Aside from BC5CDR-c where {\krissbert} already performs very well, there is a large gap (10-15 points) between top-1 and top-5 accuracy, in both self-supervised and lazy supervised settings (\autoref{fig:improve}). 
This suggests that there is much room for {\krissbert} to gain by further improving ranking. 
{\krissbert} also facilitates efficient few-shot learning, with a single example per entity yielding over 10 point gain in N2C2.
\autoref{appendix:errors} \autoref{tab:error-analysis} shows examples of common errors by {\krissbert}. They are subtle and challenging. E.g., the gold concept is expression, while {\krissbert} predicts the procedure of expression. 

\section{Conclusion}

We propose knowledge-rich self-supervised entity linking by conducting contrastive learning on mention examples generated from unlabeled text using available domain knowledge.
Experiments on seven standard biomedical entity linking datasets 
show that our proposed {\krissbert} outperforms prior state of the art by as much as 20 points in accuracy. 
Future directions include: further improving self-supervision quality; incorporating additional knowledge; applications to other domains.

\bibliography{anthology,custom}
\bibliographystyle{acl_natbib}

\appendix

\section{Additional Related Work}
\label{appendix:contrastive}

\paragraph{Zero-shot entity linking} Recent work \cite{logeswaran-etal-2019-zero} enables generalization to unseen entities by learning a cross-attention BERT model over the mention and entity contexts for candidate ranking.
\citet{gillick-etal-2019-learning,wu-etal-2020-scalable} introduce a bi-encoder that encodes the mention context and entity context separately, thus scaling to candidate generation and reducing recall loss due to mention variations. These methods, however, still require labeled information such as gold mention examples, which are not readily available in many high-value domains. This restricts their applicability to the Wikipedia domain, where labeled mentions can be gleaned from hyperlinks and entity pages. {\krissbert}, however, does not require labeled information and can learn from entity list and unlabeled text alone.
 
\paragraph{Contrastive learning} 
Contrastive learning conducts representation learning by mapping semantically similar instances to nearby points \cite{hadsell2006dimensionality}. 
Contrastive loss is often a variant of noise-contrastive estimation (NCE) that normalizes against negative (dissimilar) examples~\cite{nce}. 
A popular choice is InfoNCE~\cite{oord2018representation}, where each mini-batch samples a query instance ($q$), a few instances $k_i$'s with one positive (similar) example $k+$, and optimizes the softmax of the query's dot product with the positive example $L(q)=-\log(\exp(q\cdot k+)/\sum_i\exp(q\cdot k_i))$. In computer vision, contrastive learning is often synonymous with self-supervised learning, where ``similar'' images are generated using data augmentation techniques assumed to preserve semantics (e.g., crop, resize, recolor) \cite{wu2018unsupervised,oord2018representation,he2019moco,chen2020simple}. 
In NLP, contrastive estimation has been applied to probabilistic unsupervised learning (by approximating the partition function with a tractable neighborhood)~\cite{smith-eisner-2005-contrastive,poon-etal-2009-unsupervised}.
With the rise of neural representation, contrastive learning has also been applied to information retrieval~\cite{dssm,shen2014latent}, knowledge graph embedding~\cite{transE,distmult}, entity linking~\cite{medlinker,logeswaran-etal-2019-zero,wu-etal-2020-scalable}, question answering~\cite{karpukhin-etal-2020-dense}, typically with supervised labeled examples. 
In this paper, we apply contrastive learning to self-supervised entity linking where ``similar'' mentions are derived from unlabeled text using entity names and other domain knowledge, without requiring any labeled data.


\section{Contextual Mention Encoders}
\label{appendix:encoder}

\begin{figure}[!ht]
    \centering
    \includegraphics[width=0.32\textwidth]{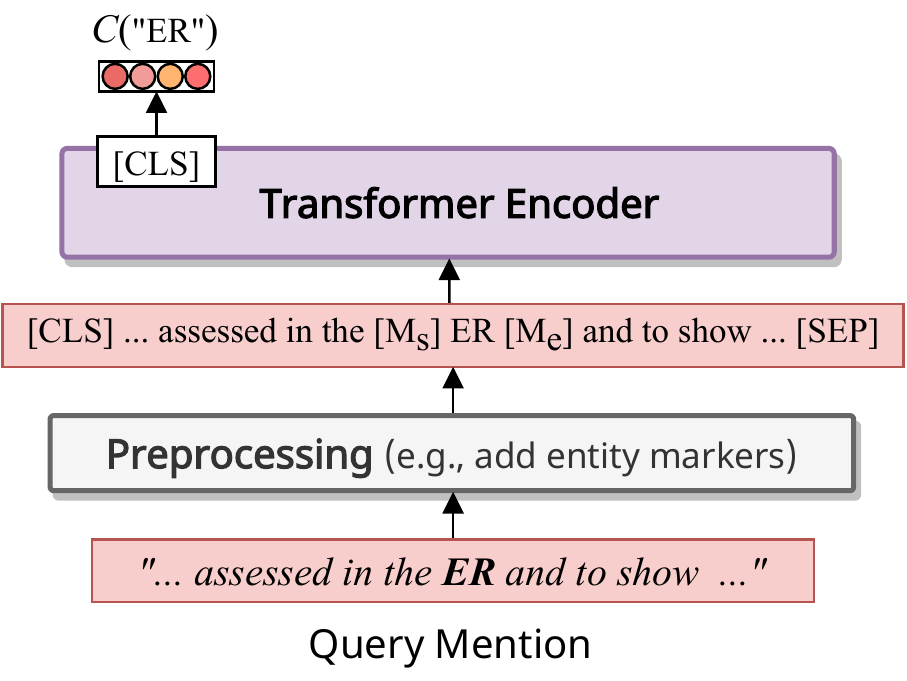}
    \caption{
    Contextual mention encoder for self-supervised entity linking.
    }
    \label{fig:encoder}
\end{figure}

\begin{figure}[!ht]
    \centering
    \includegraphics[width=0.45\textwidth]{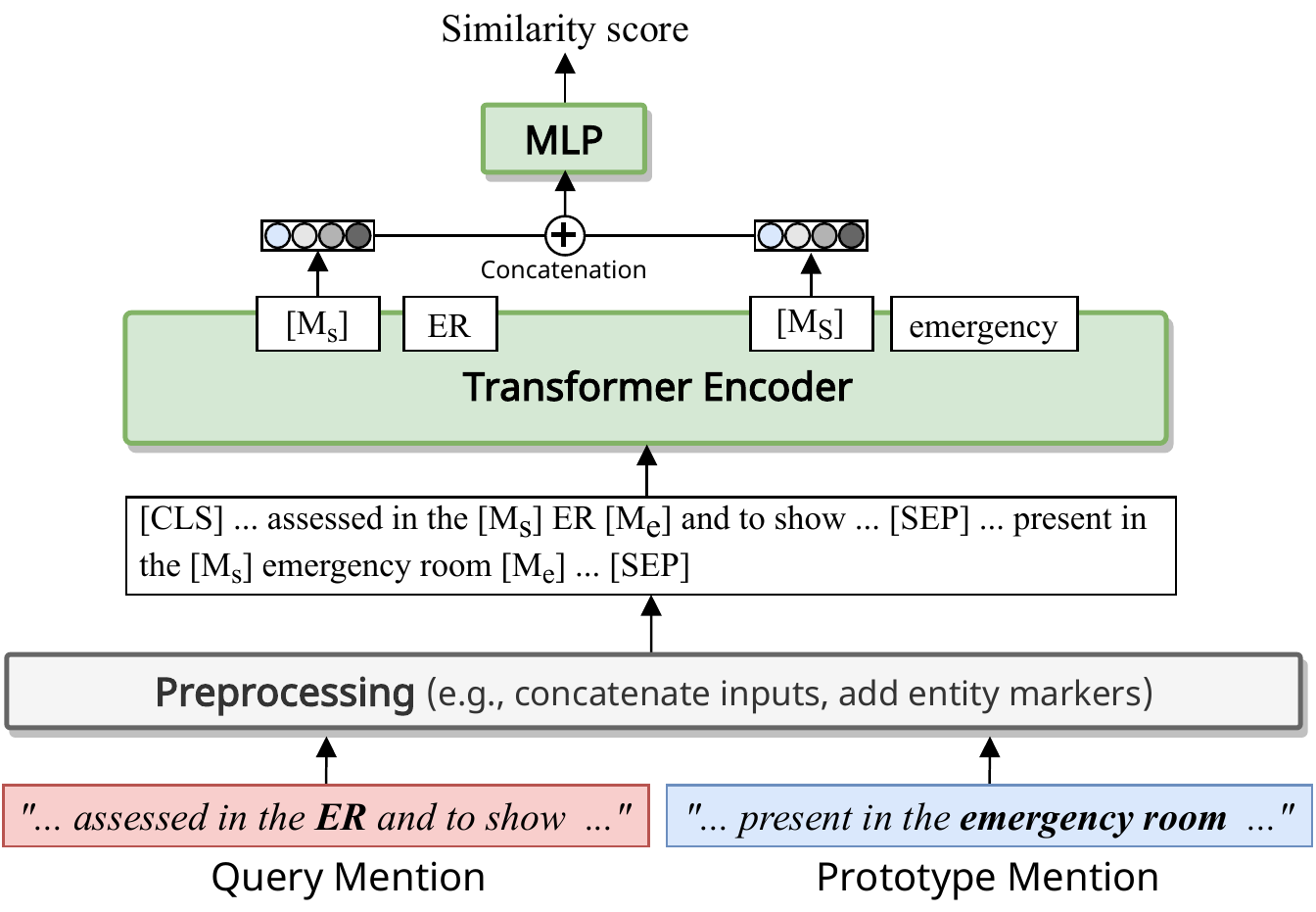}
    \caption{
    Cross-attention candidate re-ranking.
    }
    \label{fig:cross-encoder}
\end{figure}


\section{Entity Linking Datasets}
\label{appendix:dataset}

NCBI~\cite{dougan2014ncbi} contains 793 PubMed abstracts annotated with 6892 disease mentions, which are mapped to 790 unique concepts in MeSH\footnote{https://www.nlm.nih.gov/mesh/meshhome.html} or OMIM\footnote{https://omim.org/}, both part of UMLS. 
BC5CDR~\cite{BC5CDR} contains 1,500 PubMed abstracts with 5,818 annotated disease mentions (BC5CDR-d) and 4,409 chemical mentions (BC5CDR-c), which are mapped to MeSH.

ShARe~\cite{pradhan-etal-2014-semeval} contains 431 de-identified clinical reports with 17,809 disease mentions mapped to the SNOMED-CT~\cite{spackman1997snomed} subset of UMLS.

N2C2 (2019 n2c2/UMass Lowell shared task 3)~\cite{n2c2}
adds entity linking annotations to a subset of the 2010 i2b2/VA shared task dataset~\cite{uzuner20112010}. The resulting dataset contains 100 de-identified discharge summaries with 13,609 mentions (including medical problems, treatments, and tests) linked to RxNorm~\cite{liu2005rxnorm} and SNOMED-CT~\cite{spackman1997snomed} within UMLS.

MedMentions~\cite{mohan2019medmentions} (MM) is the largest publicly available dataset for biomedical entity linking, which contains 4,392 PubMed abstracts and 350,000 mentions annotated with UMLS linking. MM (st21pv) is a sub-corpus limited to 21 most common entity types.

\section{Baseline Systems}
\label{appendix:baseline}

QuickUMLS~\cite{soldaini2016quickumls} conducts entity linking by approximate matching of mentions against UMLS entity lexicon (canonical name and aliases). It serves as a representative baseline for ontology-based entity linking.

Zero-shot entity linking by reading entity descriptions~\cite{logeswaran-etal-2019-zero,wu-etal-2020-scalable} learns to encode contextual mentions against entity descriptions and attains state-of-the-art zero-shot entity linking results in the Wikipedia domain. Prior work uses gold mention examples in supervised learning. We adapt it to self-supervised learning using the self-supervised mention examples and available entity descriptions in UMLS. Prior work initializes the encoder with general-domain BERT models. To ensure head-to-head comparison, we followed {\krissbert} to use PubMedBERT~\cite{pubmedbert} instead, which yielded better results.

SapBERT~\cite{liu-etal-2021-self} learns to resolve variations in entity surface forms using synonyms in UMLS, using PubMedBERT~\cite{pubmedbert}. It ignores the mention context and returns all entities with a matching surface form. To use SapBERT for linking, we randomly select an entity when SapBERT returns multiple ones.

MedLinker~\cite{medlinker} is a strong {\em supervised entity linking} baseline that trains a BERT model on MedMentions. At test time, it augments the BERT-based prediction with approximate dictionary match for entities unseen in training.

ScispaCy~\cite{neumann-etal-2019-scispacy} provides another strong entity linking baseline that leverages labeled data in MedMentions to tune an elaborate biomedical linking system that uses TF-IDF based approximate matching and sophisticated abbreviation expansion.

\section{Error Analysis}
\label{appendix:errors}

\begin{table}[!ht]
\small
\centering
\begin{tabular}{@{}L{0.48\textwidth}@{}}
\toprule
\textbf{Mention}: ``\emph{By analysing tumor DNA from patients with sporadic t cell prolymphocytic leukemia, a rare clonal malignancy with similarities to a \underline{\textbf{mature t cell leukemia}} seen in ataxia telangiectasia ...}'' \\
\textbf{Gold entity}: \textcolor{Green}{T-Cell Leukemia (C0023492)} \\
\textbf{\krissbert~prediction}: \textcolor{NavyBlue}{T-Cell Prolymphocytic Leukemia (C2363142)} \\ \midrule
\textbf{Example}: ``\emph{The majority (81\%) of the breast ovarian cancer families were due to BRCA1, with most others (14\%) due to BRCA2. Conversely, the majority of families with \underline{\textbf{female breast cancer}} were due to BRCA2 (76\%).}'' \\
\textbf{Gold entity}: \textcolor{Green}{Breast cancer (C0006142)} \\
\textbf{\krissbert~prediction}: \textcolor{NavyBlue}{Familial cancer of breast (C0346153)} \\
\bottomrule
\end{tabular}
\caption{
Examples where {\krissbert} ``misguided'' by the context.
}
\label{tab:misguidance}
\end{table}

In \autoref{tab:misguidance}, {\krissbert} considers ``\emph{t cell prolymphocytic leukemia}'' and ``\emph{families with}'' in the context of two mentions, and predicts more specific entities than the gold ones.

\begin{table}[!t]
\small
\centering
\begin{tabular}{@{}L{0.48\textwidth}@{}}
\toprule
\textbf{Mention}: ``\emph{... NTeff cells appeared to have lower \underline{\textbf{expression}} of Foxp1  ...}'' \\
\textbf{Gold entity}: \textcolor{Green}{Protein Expression (C1171362)} \\
\textbf{\krissbert~prediction}: \textcolor{NavyBlue}{Expression Procedure (C0185117)} \\
\textbf{\krissbert~predicted prototype}: \textcolor{NavyBlue}{\emph{``... \underline{\textbf{expression}} of a myeloid differentiation antigen, Mo1 ...''}}\\ \midrule
\textbf{Mention}: ``\emph{... On admission included BUN / \underline{\textbf{creatinine}} of 33/2.1 . Sodium 141 . ...}'' \\
\textbf{Gold entity}: \textcolor{Green}{Creatinine Measurement (C0201975)} \\
\textbf{\krissbert~prediction}: \textcolor{NavyBlue}{Creatinine (C0010294)} \\
\textbf{\krissbert~predicted prototype}: \textcolor{NavyBlue}{\emph{``... Sorbent binding of urea and \underline{\textbf{creatinine}} in a Roux-Y intestinal segment. ...''}}\\ \bottomrule
\end{tabular}
\vspace{-3mm}
\caption{Examples of common errors by \krissbert.}
\label{tab:error-analysis}
\end{table}

\section{License of Scientific Artifacts}

UMLS~\cite{bodenreider2004unified} is licensed to individuals for research purposes.\footnote{\url{https://uts.nlm.nih.gov/uts/assets/LicenseAgreement.pdf}}
NCBI~\cite{dougan2014ncbi} is under the terms of the United States Copyright Act, and it is freely available to the public for use.\footnote{\url{https://huggingface.co/datasets/ncbi_disease}}
BC5CDR is freely available for the research community.\footnote{\url{https://biocreative.bioinformatics.udel.edu/tasks/biocreative-v/track-3-cdr/}}
ShARe~\cite{pradhan-etal-2014-semeval} is under The PhysioNet Credentialed Health Data License.\footnote{\url{https://physionet.org/content/shareclefehealth2014task2/view-license/1.0/}}
N2C2~\cite{n2c2} is under the Data Use and Confidentiality Agreement.\footnote{\url{https://n2c2.dbmi.hms.harvard.edu/data-use-agreement}}.
MedMentions is freely available for public use.\footnote{\url{https://github.com/chanzuckerberg/MedMentions}}
QuickUMLS~\cite{soldaini2016quickumls}, BLINK~\cite{wu-etal-2020-scalable}, SapBERT~\cite{liu-etal-2021-self}, MedLinker~\cite{medlinker}, and PubMedBERT~\cite{pubmedbert} are under the MIT License.
ScispaCy~\cite{neumann-etal-2019-scispacy} is under the Apache License 2.0.


\end{document}